\documentclass[letterpaper]{article} 
\usepackage{aaai2027}  
\usepackage[hyphens]{url}  
\usepackage{graphicx} 
\usepackage{natbib}  
\usepackage{caption} 
\usepackage{algorithm}
\usepackage{algorithmic}
\usepackage{amsmath}
\usepackage{amssymb}

\usepackage{multirow}
\usepackage{makecell}
\usepackage{newfloat}
\usepackage{listings}
\DeclareCaptionStyle{ruled}{labelfont=normalfont,labelsep=colon,strut=off} 
\floatstyle{ruled}
\newfloat{listing}{tb}{lst}{}
\floatname{listing}{Listing}

\usepackage{booktabs}

\nocopyright

\title{ReMoE: Report-Guided Mixture-of-Experts for Multimodal OCT/OCTA Anomaly Detection}

\author{
Zihan Nie\textsuperscript{\rm 1,\rm 2,\rm 6}\equalcontrib,
Qincheng Qiao\textsuperscript{\rm 4,\rm 5}\equalcontrib,
Muhao Xu\textsuperscript{\rm 1,\rm 2},
Wei Feng\textsuperscript{\rm 3,\rm 6},
Xinguo Hou\textsuperscript{\rm 4,\rm 5}\corresponding,
Weiye Song\textsuperscript{\rm 1,\rm 2}\corresponding,
Zongyuan Ge\textsuperscript{\rm 3,\rm 6}
}

\affiliations{
\textsuperscript{\rm 1}School of Mechanical Engineering, Shandong University, Jinan, China\\
\textsuperscript{\rm 2}Key Laboratory of High Efficiency and Clean Mechanical Manufacture, Shandong University, Ministry of Education, Jinan, China\\
\textsuperscript{\rm 3}Monash University, Clayton, VIC 3800, Australia\\
\textsuperscript{\rm 4}Department of Endocrinology and Metabolism, Qilu Hospital, Shandong University, Jinan 250012, China\\
\textsuperscript{\rm 5}The First Clinical Medical College, Cheeloo College of Medicine, Shandong University, Jinan 250012, China\\
\textsuperscript{\rm 6}Airdoc-Monash Research, Monash University, Clayton, VIC 3800, Australia\\
202414371@mail.sdu.edu.cn,
jugking6688@gmail.com,
ujnmhxu@hotmail.com,
wei.feng@monash.edu,
houxinguo@sdu.edu.cn,
songweiye@sdu.edu.cn,
zongyuan.ge@monash.edu
}

\begin{document}

\maketitle

\begin{abstract}
Multimodal medical anomaly detection identifies samples deviating from normal patterns, where scarce abnormal cases make normality modeling from normal data practical. In retinal Optical Coherence Tomography (OCT) and OCT Angiography (OCTA) anomaly detection, existing unsupervised methods rely on visual feature distributions, reconstruction residuals, or encoder-decoder discrepancies, making anomaly scores depend on appearance-level deviations, while multimodal normality also contains semantic organization described in normal medical reports. To this end, we propose Report-Guided Mixture-of-Experts (ReMoE), which distills normal report semantics into an image-to-text prior student, builds modality-aware priors, and uses Report-Guided Modality Modulation (RMM) to modulate features through mixture-of-experts routing. Experiments on a private OCT/OCTA dataset with paired normal reports and a public OCTA500-3MM setting using a fixed normal report demonstrate state-of-the-art performance.\end{abstract}

\section{Introduction}

Medical anomaly detection aims to identify samples that deviate from normal patterns, with unsupervised formulations commonly adopted because abnormal cases are often scarce, diverse, and incompletely annotated~\cite{bergmann2019mvtec,roth2022towards,deng2022anomaly}. In this setting, models learn normality from normal data and detect deviations at test time, making normal-state representation central to detection performance. When multiple imaging modalities are available, complementary structural, functional, and spatial information can enrich normality modeling~\cite{basu2024systematic,xu2026towards}, while such complementary observations also make anomaly detection more challenging because normal states are jointly characterized by modality-specific visual patterns and cross-modality relationships. In retinal multimodal imaging, Optical Coherence Tomography (OCT) depicts retinal layer structures, tissue continuity, and macular morphology, while OCT Angiography (OCTA) captures vascular patterns, blood-flow distribution, and perfusion status~\cite{huang1991optical,spaide2018optical,li2024octa,he2025fusing}. Effective anomaly detection in this setting therefore requires representations that capture multimodal normality beyond image fusion.
\begin{figure}[t]
\centering
\includegraphics[width=\columnwidth]{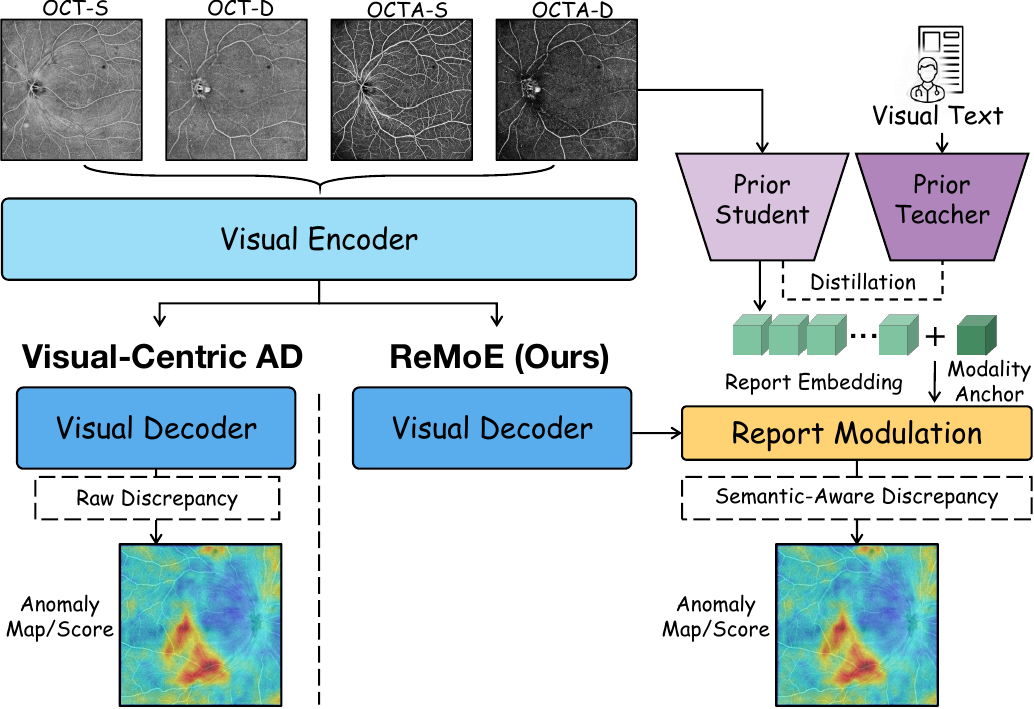}
\caption{Visual-centric anomaly detection versus report-guided ReMoE.}
\label{fig:teaser}
\end{figure}

To learn such normality representations, existing anomaly detection methods mainly model normal visual patterns and measure test-time deviations through reconstruction residuals~\cite{zavrtanik2021draem}, feature distributions~\cite{defard2021padim}, or encoder-decoder feature discrepancies~\cite{deng2022anomaly,nie2026cfr}. These visual centric paradigms make anomaly scores largely depend on raw visual deviations, while retinal reports describe structural and angiographic findings at a semantic level. This motivates the use of image-text learning to introduce useful semantic signals, as demonstrated by medical vision-language studies~\cite{zhang2022contrastive,huang2021gloria}. In multimodal OCT/OCTA anomaly detection, transforming report semantics into modality aware priors allows semantic guidance to align with structural, vascular, and layer specific cues during feature discrepancy modeling. Figure~\ref{fig:teaser} illustrates this motivation by comparing visual centric anomaly detection with the proposed report guided semantic aware discrepancy modeling.

To address this challenge, we propose Report-Guided Mixture-of-Experts (ReMoE) for multimodal anomaly detection, a framework that introduces report-derived semantic priors into the feature discrepancy modeling process of an encoder-decoder anomaly detector. Specifically, ReMoE extracts normal report embeddings with a frozen text encoder and uses them as distillation targets for an image-to-text prior student, so that report-related normal semantics can be predicted from multimodal images at test time. Combined with modality anchors, these pseudo-text representations form modality-aware priors, which are used by Report-Guided Modality Modulation (RMM) to route visual experts and modulate encoder and decoder features before discrepancy computation, converting raw visual feature discrepancies into semantic-aware discrepancies and allowing anomaly scores to reflect deviations from visual normality under report-guided modality modulation.

We evaluate ReMoE on a private OCT/OCTA dataset with paired reports and the public OCTA500-3MM setting. Results show that ReMoE outperforms existing visual and multimodal anomaly detection methods. The main contributions are:
\begin{itemize}
    \item We propose ReMoE, a report-guided mixture-of-experts framework that distills normal report semantics into image-conditioned pseudo-text representations for multimodal anomaly detection.
    \item We design RMM to use modality-aware priors for expert routing and encoder-decoder feature modulation, enabling semantic-aware anomaly scoring.
    \item Extensive experiments on private and public OCT/OCTA settings demonstrate consistent improvements over existing unimodal and multimodal anomaly detection methods.
\end{itemize}

\section{Related Work}

\subsection{Medical Anomaly Detection}

Medical anomaly detection methods identify anomalies from deviations between test samples and learned normal representations. Existing methods derive anomaly evidence from reconstruction residuals~\cite{zavrtanik2021draem,wyatt2022anoddpm}, feature distributions and memory banks~\cite{defard2021padim,roth2022towards}, multiscale flow likelihoods~\cite{zhou2024msflow}, teacher and student matching~\cite{wang2021student}, discriminative feature learning~\cite{liu2023simplenet}, or encoder and decoder feature discrepancies~\cite{deng2022anomaly,nie2026cfr}. Recent methods improve reverse distillation with expert teacher student networks~\cite{liu2025unlocking} and exploit Transformer based foundation features for multi class unsupervised learning~\cite{guo2025dinomaly}. In medical scenarios, anomaly detection has been extended to few shot calibration~\cite{nie2026few}, cross modality PET/CT distillation~\cite{xu2026towards}, and multimodal retinal anomaly detection~\cite{li2025adapting}, while recent multimodal frameworks have further explored flexible handling of incomplete MRI sequences~\cite{wu2025anyad} and unified multimodal multiclass anomaly detection~\cite{zhao2026unimmad}. Extending multimodal normality modeling beyond visual integration, ReMoE incorporates image conditioned report semantics through encoder and decoder feature discrepancy modulation.

\subsection{Medical Vision-Language Learning}

Medical vision-language learning introduces textual semantics into medical visual representation learning. Image-report contrastive learning uses paired medical images and reports to learn transferable representations~\cite{zhang2022contrastive}, while global and local alignment further associates image regions with report phrases for label-efficient recognition~\cite{huang2021gloria}. Medical image-text learning has also been studied with unpaired medical images and text~\cite{wang2022medclip}, as well as biomedical text semantics for broader vision and language processing~\cite{boecking2022making}. In ophthalmology, multimodal retinal images and clinical text have been jointly modeled for retinal analysis~\cite{shi2025multimodal}, while text-guided medical anomaly detection has used clinical text prototypes to provide normal semantic references under normal-only training~\cite{lin2026ufmads}. Building on these advances, ReMoE uses paired normal reports to supervise an image-to-text prior student that predicts image-conditioned pseudo-text representations for report-modulated feature discrepancy modeling.

\section{Method}

\subsection{Framework Overview}

\begin{figure*}[t]
    \centering
    \includegraphics[width=\textwidth]{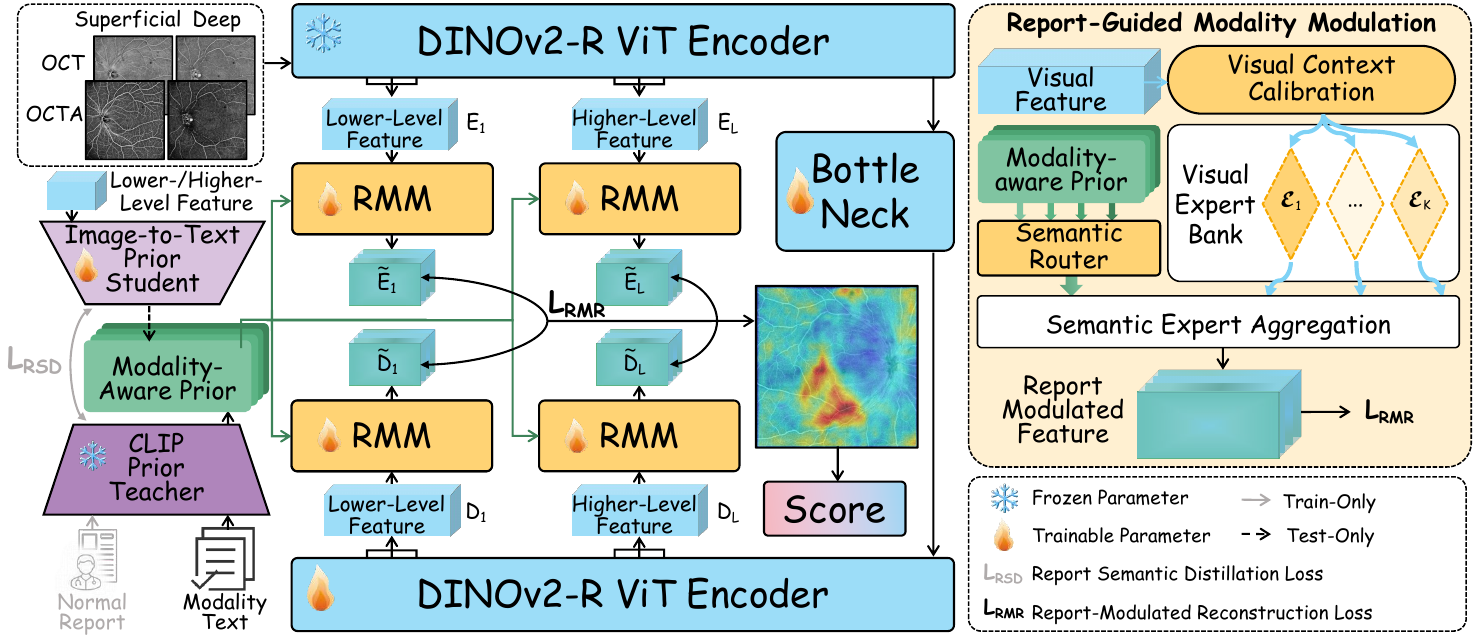}
    \caption{Overview of ReMoE. A lightweight image-to-text prior student is distilled from a frozen CLIP prior teacher and combines the predicted pseudo-text representation with modality anchors to construct modality-aware priors. These priors guide RMM to modulate paired encoder and decoder features before anomaly discrepancy computation.}
    \label{fig:framework}
\end{figure*}

As illustrated in Figure~\ref{fig:framework}, ReMoE takes four single-channel retinal inputs from the same case, including superficial and deep OCT and OCTA, and concatenates them along the channel dimension to form $\mathbf{X}\in\mathbb{R}^{B\times M\times H\times W}$, where \(B\) is the batch size, \(M\) is the number of inputs (\(M=4\) in the full setting), and \(H\) and \(W\) are the spatial dimensions. We average its pretrained input projection weights across the three channels, repeat the averaged weights \(M\) times, and rescale them by \(3/M\),  which remains frozen with the encoder.

The visual backbone consists of a frozen DINOv2-R ViT encoder~\cite{oquab2023dinov2,darcet2024vision}, a trainable bottleneck, and the trainable ViT decoder. The encoder extracts intermediate visual representations from \(\mathbf{X}\). Following Dinomaly~\cite{guo2025dinomaly}, intermediate Transformer features are aggregated into multiple feature levels through cross-layer averaging, producing encoder features $\mathbf{E}_l\in\mathbb{R}^{B\times C\times h\times w}$, where \(l\in\{1,\ldots,L\}\) indexes a feature level, \(L\) is the total number of feature levels, \(C\) is the channel dimension, and \(h\) and \(w\) are the spatial dimensions of the feature maps. The encoded representations are subsequently passed through the bottleneck and ViT decoder, and cross-layer aggregation is applied to obtain the corresponding decoder features $\mathbf{D}_l\in\mathbb{R}^{B\times C\times h\times w}$.

At each feature level, a lightweight image-to-text prior student predicts a pseudo-text representation, which is combined with modality anchors to construct modality-aware priors, collectively denoted by $\mathbf{P}_l^q$. RMM uses these priors to modulate the paired encoder and decoder features:
\begin{equation}
    \widetilde{\mathbf{V}}_l^q
    =
    \operatorname{RMM}_{l}^{q}
    \left(
    \mathbf{V}_l^q,
    \mathbf{P}_l^q
    \right),
    \qquad q\in\{E,D\},
\end{equation}
where $\mathbf{V}_l^E=\mathbf{E}_l$, $\mathbf{V}_l^D=\mathbf{D}_l$, and $\widetilde{\mathbf{V}}_l^E=\widetilde{\mathbf{E}}_l$, $\widetilde{\mathbf{V}}_l^D=\widetilde{\mathbf{D}}_l$. The report semantic distillation, modality-aware prior construction and the internal design of RMM are described in the following subsections.

The model learns normal visual patterns by minimizing the discrepancy between report-modulated encoder and decoder features at corresponding levels. We define the report-modulated reconstruction loss as
\begin{equation}
    \mathcal{L}_{\mathrm{RMR}}
    =
    \frac{1}{BL}
    \sum_{b=1}^{B}\sum_{l=1}^{L}
    \left[
    1-
    \operatorname{cos}
    \left(
    \operatorname{vec}(\widetilde{\mathbf{E}}_{b,l}),
    \operatorname{vec}(\widetilde{\mathbf{D}}_{b,l})
    \right)
    \right],
\end{equation}
where \(b\) indexes a sample in the batch, \(\operatorname{vec}(\cdot)\) vectorizes a feature map, and \(\operatorname{cos}(\cdot,\cdot)\) denotes cosine similarity. The complete training objective is
\begin{equation}
    \mathcal{L}_{\mathrm{total}}
    =
    \mathcal{L}_{\mathrm{RMR}}
    +
    \lambda_{\mathrm{RSD}}\mathcal{L}_{\mathrm{RSD}},
\end{equation}
where \(\lambda_{\mathrm{RSD}}\) controls the contribution of report semantic distillation loss \(\mathcal{L}_{\mathrm{RSD}}\), which is detailed in the following subsection.

During inference, the pixel-level anomaly map is obtained by averaging the upsampled channel-wise cosine discrepancies across feature levels:
\begin{equation}
    \mathbf{A}
    =
    \frac{1}{L}
    \sum_{l=1}^{L}
    \operatorname{Up}
    \left(
    1-
    \operatorname{cos}_{c}
    \left(
    \widetilde{\mathbf{E}}_l,
    \widetilde{\mathbf{D}}_l
    \right)
    \right),
\end{equation}
where $\operatorname{cos}_{c}(\cdot,\cdot)$ denotes channel-wise cosine similarity at each spatial position, and $\operatorname{Up}(\cdot)$ upsamples the discrepancy map to the input resolution using bilinear interpolation. The image-level anomaly score is calculated by averaging the highest-response proportion $\rho$ of pixels:
\begin{equation}
    s(\mathbf{X})
    =
    \frac{1}{|\Omega_{\rho}|}
    \sum_{(i,j)\in\Omega_{\rho}}
    \mathbf{A}(i,j),
\end{equation}
where $\Omega_{\rho}$ is the set containing the top-$\rho$ pixels in $\mathbf{A}$, and $|\Omega_{\rho}|$ is its cardinality.

\subsection{Report-Guided Semantic Prior Learning}
\label{sec:semantic_prior}

During training, normal reports guide a lightweight image-to-text prior student to learn a visual-to-text semantic mapping, enabling it to predict pseudo-text representations from visual features during inference and further construct modality-aware priors for visual modulation.

Given the normal report $r_b$ of the $b$-th training sample, the frozen CLIP~\cite{radford2021learning} prior teacher $f_{\mathrm{text}}$ encodes it into a text embedding:
\begin{equation}
    \mathbf{t}_b
    =
    \operatorname{Norm}
    \big(
    f_{\mathrm{text}}(r_b)
    \big),
    \qquad
    \mathbf{t}_b\in\mathbb{R}^{d},
\end{equation}
where $d$ is the text embedding dimension, and $\operatorname{Norm}(\cdot)$ denotes $L_2$ normalization. The resulting embedding provides semantic supervision for learning pseudo-text representations.

Given the visual feature $\mathbf{F}_l^q\in\mathbb{R}^{B\times C\times h\times w}$ used by the prior student, where $\mathbf{F}_l^q$ corresponds to the encoder feature $\mathbf{E}_l$ or decoder feature $\mathbf{D}_l$ before report modulation, the lightweight image-to-text prior student applies global average pooling followed by a linear projection to obtain a pseudo-text representation:
\begin{equation}
    \mathbf{s}_{b,l}^{q}
    =
    \operatorname{Norm}
    \left(
    \mathbf{W}_s^{q}
    \operatorname{GAP}
    \left(
    \mathbf{F}_{b,l}^{q}
    \right)
    \right),
    \qquad
    \mathbf{s}_{b,l}^{q}
    \in
    \mathbb{R}^{d},
\end{equation}
where $\operatorname{GAP}(\cdot)$ denotes global average pooling, $\mathbf{W}_s^{q}$ is the learnable projection on network side $q$, and student parameters are shared across feature levels within the same network side.

To distill normal report semantics into the student, we define the report semantic distillation loss as
\begin{equation}
    \mathcal{L}_{\mathrm{RSD}}
    =
    \frac{1}{2BL}
    \sum_{q\in\{E,D\}}
    \sum_{b=1}^{B}
    \sum_{l=1}^{L}
    \left[
    1-
    \left\langle
    \mathbf{s}_{b,l}^{q},
    \mathbf{t}_b
    \right\rangle
    \right],
\end{equation}
where $\langle\cdot,\cdot\rangle$ denotes the inner product. This alignment establishes a normal semantic reference for the pseudo-text representations, so deviations in test images may lead to different semantic conditions for RMM.

To disentangle shared normal semantics into modality-specific semantic directions, we define a modality text for each input modality and obtain the corresponding modality anchors using the frozen CLIP text encoder:
\begin{equation}
    \mathbf{a}_m
    =
    \operatorname{Norm}
    \big(
    f_{\mathrm{text}}(u_m)
    \big),
    \qquad
    m\in\{1,\ldots,M\},
\end{equation}
where $u_m$ denotes the text description of the $m$-th modality, and $\mathbf{a}_m\in\mathbb{R}^{d}$ is the corresponding modality anchor.

The pseudo-text representation is first combined with the corresponding modality anchor:
\begin{equation}
    \mathbf{r}_{b,l,m}^{q}
    =
    \operatorname{Norm}
    \left(
    \mathbf{s}_{b,l}^{q}
    +
    \mathbf{a}_m
    \right).
\end{equation}

The combined semantic representation is then processed by two lightweight projection heads. The first predicts an image-conditioned semantic component:
\begin{equation}
\mathbf{p}_{b,l,m}^{q,\mathrm{stu}}
=
\operatorname{Norm}
\left(
\phi_s^{q}
\left(
\mathbf{r}_{b,l,m}^{q}
\right)
\right),
\end{equation}
and the second predicts the contribution coefficient of
$\mathbf{p}_{b,l,m}^{q,\mathrm{stu}}$:
\begin{equation}
\gamma_{b,l,m}^{q}
=
\sigma
\left(
\left[
g_{\gamma}^{q}
\left(
\mathbf{r}_{b,l,m}^{q}
\right)
\right]_m
\right),
\end{equation}
where $\phi_s^{q}:\mathbb{R}^{d}\rightarrow\mathbb{R}^{C}$ denotes the semantic projection, $g_{\gamma}^{q}:\mathbb{R}^{d}\rightarrow\mathbb{R}^{M}$ denotes the contribution predictor, $\sigma(\cdot)$ is the sigmoid function, and $[\cdot]_m$ selects the $m$-th element.

Meanwhile, to preserve the intrinsic semantic direction of each modality, the modality anchor is transformed into a modality-specific semantic component:
\begin{equation}
    \mathbf{p}_{m}^{q,\mathrm{anc}}
    =
    \phi_a^{q}
    \left(
    \mathbf{a}_m
    \right),
    \qquad
    \phi_a^{q}:\mathbb{R}^{d}\rightarrow\mathbb{R}^{C}.
\end{equation}

Finally, the image-conditioned semantic component adaptively refines the modality-specific prior to construct the modality-aware prior:
\begin{equation}
\label{eq:prior_fusion}
    \mathbf{p}_{b,l,m}^{q}
    =
    \mathbf{p}_{m}^{q,\mathrm{anc}}
    +
    \tanh(\alpha^q)
    \gamma_{b,l,m}^{q}
    \mathbf{p}_{b,l,m}^{q,\mathrm{stu}},
\end{equation}
where $\alpha^q$ is a learnable scalar controlling the overall contribution of the image-conditioned semantic component. The resulting priors $\mathbf{p}_{b,l,m}^{q}$ constitute $\mathbf{P}_l^q$, which is subsequently used as the semantic input to RMM.

\subsection{Report-Guided Modality Modulation}
\label{sec:rmm}
RMM uses modality-aware priors to generate expert-routing weights and modulate paired encoder and decoder features before discrepancy computation. Let $\mathbf{V}_l^E=\mathbf{E}_l$ and $\mathbf{V}_l^D=\mathbf{D}_l$ denote the encoder and decoder features, respectively, with corresponding priors $\mathbf{P}_l^q\in\mathbb{R}^{B\times M\times C}$.

RMM first performs multi-scale visual context modeling on the input visual feature. Specifically, $\mathbf{V}_l^q$ is processed by parallel depthwise separable convolution branches, forming a multi-scale context transformation that is applied in a residual form to obtain the calibrated visual representation:
\begin{equation}
    \mathbf{H}_l^q
    =
    \mathbf{V}_l^q
    +
    \mathcal{C}_l^q
    \left(
    \mathbf{V}_l^q
    \right),
\end{equation}
where $\mathcal{C}_l^q(\cdot)$ denotes the context transformation composed of multi-scale depthwise separable convolution branches, and $\mathbf{H}_l^q\in\mathbb{R}^{B\times C\times h\times w}$ is the calibrated visual feature.

RMM then generates semantics-conditioned expert routing for different modalities based on the modality-aware priors. For the $b$-th sample and the $m$-th modality, the prior $\mathbf{p}_{b,l,m}^{q}$ is projected into the visual feature space and additively fused with the calibrated visual feature:
\begin{equation}
    \mathbf{G}_{b,l,m}^{q}
    =
    \mathbf{H}_{b,l}^{q}
    +
    \psi_l^q
    \left(
    \mathbf{p}_{b,l,m}^{q}
    \right),
\end{equation}
where $\psi_l^q(\cdot)$ denotes the prior-to-visual projection, and $\mathbf{G}_{b,l,m}^{q}\in\mathbb{R}^{C\times h\times w}$ is the routing feature formed by visual context and modality semantics. The semantic router globally pools this routing feature and predicts spatially shared expert weights for each modality:
\begin{equation}
    \mathbf{w}_{b,l,m}^{q}
    =
    \operatorname{Softmax}
    \left(
    \mathcal{R}_l^q
    \left(
    \mathbf{G}_{b,l,m}^{q}
    \right)
    \right),
    \qquad
    \mathbf{w}_{b,l,m}^{q}\in\mathbb{R}^{K},
\end{equation}
where $\mathcal{R}_l^q(\cdot)$ denotes the semantic router, $K$ is the number of visual experts, and $\mathbf{w}_{b,l,m}^{q}(k)$ denotes the weight assigned by the $m$-th input condition to the $k$-th expert.

Meanwhile, RMM uses a set of independently parameterized visual experts to transform the calibrated visual feature. The response of the $k$-th visual expert is given by
\begin{equation}
    \mathbf{O}_{b,l,k}^{q}
    =
    \mathcal{E}_{l,k}^{q}
    \left(
    \mathbf{H}_{b,l}^{q}
    \right),
    \qquad
    k=1,\ldots,K,
\end{equation}
where $\mathbf{O}_{b,l,k}^{q}\in\mathbb{R}^{C\times h\times w}$ denotes the response of the $k$-th expert.
Different experts use distinct convolutional receptive fields, and their responses are combined according to modality-conditioned routing weights.

Based on the routing weights and expert responses, RMM jointly aggregates over modalities and experts to obtain the semantics-modulated response:
\begin{equation}
    \mathbf{U}_{b,l}^{q}
    =
    \frac{1}{M}
    \sum_{m=1}^{M}
    \sum_{k=1}^{K}
    \mathbf{w}_{b,l,m}^{q}(k)
    \mathbf{O}_{b,l,k}^{q}.
\end{equation}
Finally, the semantics-modulated response is fused with the calibrated visual feature in a residual form, followed by an output projection to obtain the report-modulated feature:
\begin{equation}
    \widetilde{\mathbf{V}}_{b,l}^{q}
    =
    \eta_l^q
    \left(
    \mathbf{H}_{b,l}^{q}
    +
    \mathbf{U}_{b,l}^{q}
    \right),
\end{equation}
where $\eta_l^q(\cdot)$ denotes the output projection, and $\widetilde{\mathbf{V}}_{b,l}^{q}$ is the report-modulated feature, corresponding to $\widetilde{\mathbf{E}}_{b,l}$ on the encoder side and $\widetilde{\mathbf{D}}_{b,l}$ on the decoder side.

\begin{table*}[t]
\centering
\small
\setlength{\tabcolsep}{1.3pt}
\renewcommand{\arraystretch}{1.08}
\begin{tabular*}{\textwidth}{@{\extracolsep{\fill}}cccccccc@{}}
\toprule
\multirow{2}{*}{Category} & \multirow{2}{*}{Method} & \multirow{2}{*}{Inputs} & \multirow{2}{*}{Protocol}
& \multicolumn{2}{c}{Private OCT/OCTA}
& \multicolumn{2}{c}{OCTA500-3MM} \\
\cmidrule{5-6}\cmidrule{7-8}
 &  &  &  & I-AUROC & I-AP & I-AUROC & I-AP \\
\midrule
\multirowcell{6}{Visual only\\unimodal}
& MSFlow & 1 & Avg. 4 inputs
& $0.6532\pm0.0067$ & $0.7969\pm0.0048$
& $0.8432\pm0.0074$ & $0.8568\pm0.0064$ \\
& URD & 1 & Avg. 4 inputs
& $0.6773\pm0.0049$ & $0.8091\pm0.0050$
& $0.8452\pm0.0090$ & $0.8454\pm0.0091$ \\
& Dinomaly & 1 & Avg. 4 inputs
& $0.7175\pm0.0015$ & $0.8511\pm0.0010$
& $0.8242\pm0.0061$ & $0.8302\pm0.0097$ \\
& RD4AD & 1 & Avg. 4 inputs
& $0.6903\pm0.0013$ & $0.7746\pm0.0129$
& $0.8222\pm0.0017$ & $0.8316\pm0.0024$ \\
& PatchCore & 1 & Avg. 4 inputs
& $0.6776\pm0.0009$ & $0.7324\pm0.0010$
& $0.6973\pm0.0038$ & $0.7367\pm0.0043$ \\
& SimpleNet & 1 & Avg. 4 inputs
& $0.7258\pm0.0053$ & $0.8386\pm0.0029$
& $0.7901\pm0.0042$ & $0.8142\pm0.0081$ \\
\midrule
\multirowcell{4}{Visual only\\multimodal}
& MMRAD & 2 & Avg. 2 pairs
& $0.5705\pm0.0358$ & $0.7392\pm0.0225$
& $0.6454\pm0.0702$ & $0.7007\pm0.0670$ \\
& TRMUAD & 2 & Avg. 2 pairs
& $0.7016\pm0.0027$ & $0.8236\pm0.0066$
& $0.8519\pm0.0046$ & $0.8602\pm0.0037$ \\
& AnyAD & 4 & All 4 inputs
& $0.7800\pm0.0016$ & $0.8925\pm0.0013$
& $0.8589\pm0.0015$ & $0.8898\pm0.0042$ \\
& UniMMAD & 4 & All 4 inputs
& $0.7015\pm0.0019$ & $0.8330\pm0.0015$
& $0.8024\pm0.0032$ & $0.8258\pm0.0040$ \\
\midrule
\multirowcell{2}{Ours}
& ReMoE-2I & 2 & Avg. 2 pairs
& $0.7994\pm0.0041$ & $0.8908\pm0.0032$
& $0.8612\pm0.0489$ & $0.8792\pm0.0450$ \\
& ReMoE & 4 & All 4 inputs
& $\mathbf{0.8682\pm0.0033}$ & $\mathbf{0.9358\pm0.0020}$
& $\mathbf{0.8995\pm0.0232}$ & $\mathbf{0.9130\pm0.0244}$ \\
\bottomrule
\end{tabular*}
\caption{Comparison with existing methods on private and public OCT/OCTA settings.}
\label{tab:main_results}
\end{table*}

\section{Experiments}

\subsection{Datasets and Evaluation Protocol}

We evaluate ReMoE on a private OCT/OCTA dataset and an OCTA500-3MM setting constructed from the public OCTA500 dataset~\cite{li2024octa}. 

\textbf{Private OCT/OCTA dataset.}
The private dataset contains paired multimodal retinal images and clinical reports. Each case includes superficial OCT, deep OCT, superficial OCTA, and deep OCTA with image-level classification labels. The superficial slab is defined from the Internal Limiting Membrane (ILM) to the Inner Plexiform Layer (IPL), and the deep slab is defined from IPL to the Outer Plexiform Layer (OPL). The dataset is randomly split at the patient level, with all cases from the same patient assigned to the same subset, resulting in 374 normal training cases and a test set with 1,436 normal and 2,981 abnormal cases. Image-level labels and paired reports are derived from de-identified clinical records; further details are provided in the Supplementary Material.

\textbf{OCTA500-3MM.}
For OCTA500-3MM, we use the 3-mm subset and organize four OCT/OCTA inputs from different modality and slab combinations: OCT ILM-OPL, OCT OPL-BM, OCTA ILM-OPL, and OCTA OPL-BM, where BM denotes Bruch's Membrane. Since OCTA500-3MM does not provide paired clinical reports, all normal training samples share a fixed normal report generated by GPT~\cite{singh2026openaigpt5card}, which serves as a common normal semantic reference for this public setting. We train on 120 normal samples and evaluate on 40 normal and 40 abnormal samples.

\textbf{Evaluation protocol.}
All baselines are reproduced following their original configurations. We report the average over four single-input results for unimodal baselines, the average over two matched OCT/OCTA pairs for two-input multimodal baselines, and direct four-input results for four-input methods. All experiments are repeated with five random seeds, and we report mean $\pm$ std using Image-level Area Under the Receiver Operating Characteristic Curve (I-AUROC) and Image-level Average Precision (I-AP).

De-identified report examples, the fixed normal report used for OCTA500-3MM, and the modality anchor texts are provided in the Supplementary Material.

\subsection{Implementation Details}

Input images are resized to $448\times448$ and center-cropped to $392\times392$. The visual backbone is a frozen DINOv2-R ViT-S/14~\cite{oquab2023dinov2,darcet2024vision}. The CLIP prior teacher is a frozen CLIP ViT-B/32 text encoder~\cite{radford2021learning}. 

We train the model with StableAdamW for 10,000 iterations. The initial and final learning rates are $2\times10^{-3}$ and $2\times10^{-4}$, respectively, with weight decay $10^{-4}$, betas $(0.9,0.999)$, 100 warmup iterations, and cosine decay. The training and testing batch sizes are 8 and 32. Gradient clipping is applied with a maximum norm of 0.1. We use $\lambda_{\mathrm{RSD}}=0.15$, and the learnable report scale parameter is initialized to 0 and optimized jointly with the model. RMM uses $K=3$ visual experts with $1\times1$, $3\times3$, and $5\times5$ depthwise convolution branches. The routing temperature is 1.0, and the image-level score averages the top-$\rho$ pixel responses with $\rho=0.01$. All experiments are conducted on an NVIDIA A100 40GB GPU. Efficiency and resource comparisons under the case-level inference protocol are provided in the Supplementary Material.

\section{Results and Analysis}

\begin{figure*}[t!]
\centering
\includegraphics[width=\textwidth]{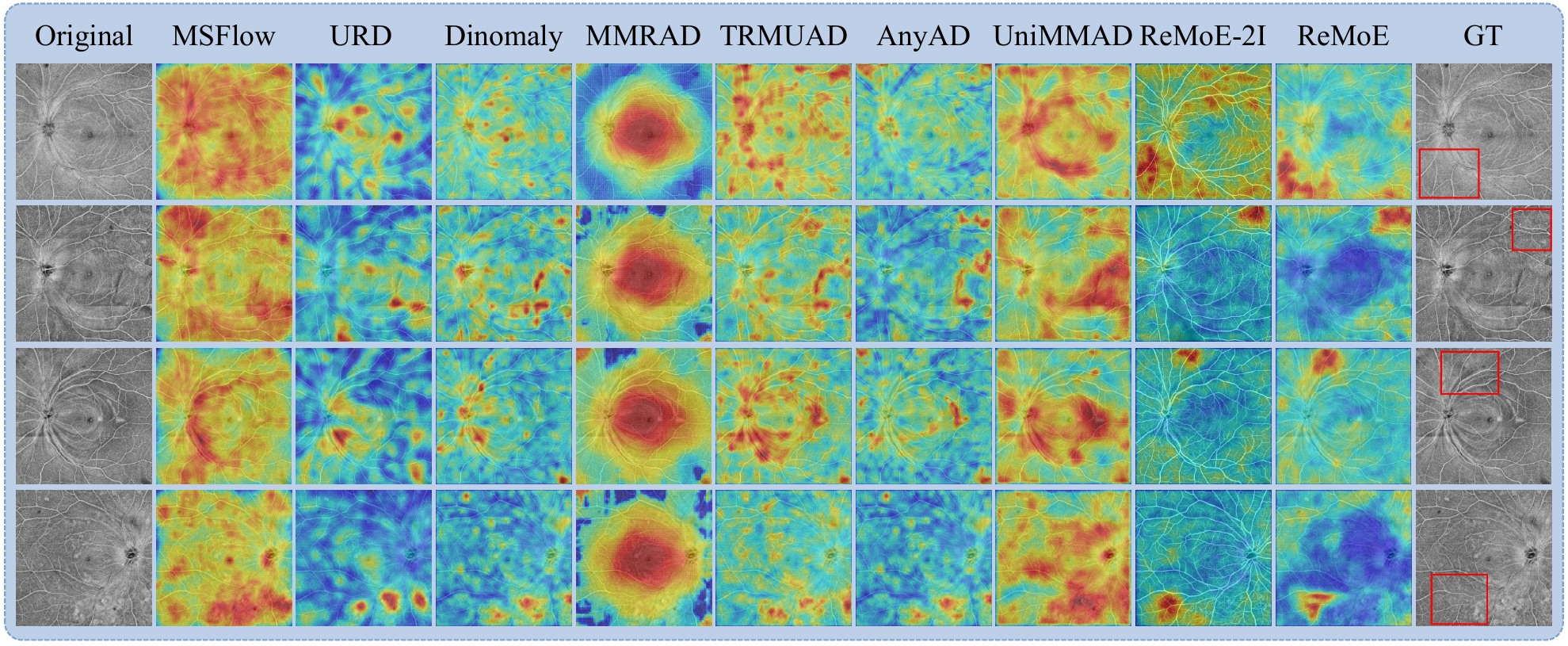}
\caption{Qualitative anomaly maps on the private OCT/OCTA dataset. The red boxes indicate abnormal regions annotated by an ophthalmologist according to the corresponding clinical reports.}
\label{fig:qualitative}
\end{figure*}

\subsection{Comparison with Existing Methods}
\label{sec:comparison}
In Table~\ref{tab:main_results}, we compare ReMoE with existing methods on private and public OCT/OCTA settings. ReMoE achieves state-of-the-art performance in both settings. On the private dataset, it reaches an AUROC of 0.8682 and an AP of 0.9358, outperforming the strongest four-input visual-only baseline, AnyAD, by 8.82 and 4.33 percentage points, respectively. On OCTA500-3MM, ReMoE improves over AnyAD by 4.06 percentage points in AUROC and 2.32 percentage points in AP.

In the two-input setting, the two-input ReMoE variant (ReMoE-2I) consistently outperforms the two-input multimodal baselines and remains competitive with four-input visual-only methods. This indicates that report-derived priors can organize the shared representation according to modality conditions before discrepancy computation, allowing paired structural and vascular cues to be used more effectively.

Using all four OCT/OCTA inputs further improves ReMoE-2I by 6.88 and 4.50 percentage points on the private dataset, and by 3.83 and 3.38 percentage points on OCTA500-3MM, in terms of AUROC and AP, respectively. The further gains show that complete structural, vascular, and layer information provides richer evidence for modality-aware routing, enabling RMM to refine the shared representation and better distinguish discrepancies that disrupt normal cross-modality relationships. Detailed single-input, matched-pair, and seed-wise results are reported in the Supplementary Material.

In Figure~\ref{fig:qualitative}, we show qualitative anomaly maps on the private OCT/OCTA dataset, where red boxes indicate abnormal regions described in the clinical reports. Compared with other methods, ReMoE produces stronger and more spatially coherent responses that are better localized around report-indicated regions, while showing fewer background activations, suggesting that normal report priors help suppress less relevant discrepancies and highlight abnormal visual patterns. Compared with ReMoE-2I, ReMoE further improves response strength and localization consistency, as the additional modalities and slabs provide more complete visual evidence for anomaly detection.

\subsection{Ablation Studies}
Ablations in this section are conducted on the private OCT/OCTA dataset with reported values averaged over five random seeds, and key OCTA500-3MM ablations are provided in the Supplementary Material.

\textbf{Component ablation.}
Table~\ref{tab:component_ablation} analyzes key components using the four-input DINOv2 encoder-decoder discrepancy model without RMM as the baseline. Visual only RMM routes experts from image features and yields a modest improvement. Anchor only RMM performs better by introducing modality anchors as routing conditions, showing that modality and slab semantics help organize the shared representation before discrepancy computation.
\begin{table}[h]
\centering
\small
\setlength{\tabcolsep}{0.6pt}
\begin{tabular*}{\columnwidth}{@{\extracolsep{\fill}}lcccccc@{}}
\toprule
Variant & Prior & Stu. & RMM & Anc. & I-AUROC & I-AP \\
\midrule
Baseline
& $-$ & $-$ & $-$ & $-$ & $0.7666$ & $0.8831$ \\
+ Visual only RMM
& $-$ & $-$ & $\checkmark$ & $-$ & $0.7732$ & $0.8886$ \\
+ Anchor only RMM
& $-$ & $-$ & $\checkmark$ & $\checkmark$ & $0.7801$ & $0.8937$ \\
+ Random text prior
& Rand. & $-$ & $\checkmark$ & $-$ & $0.7764$ & $0.8901$ \\
+ Shuffled report prior
& Shuf. & $-$ & $\checkmark$ & $-$ & $0.7918$ & $0.8996$ \\
+ Fixed report prior
& Fix. & $-$ & $\checkmark$ & $-$ & $0.8007$ & $0.9035$ \\
+ Student only RMM
& Rep. & $\checkmark$ & $\checkmark$ & $-$ & $0.8233$ & $0.9074$ \\
ReMoE
& Rep. & $\checkmark$ & $\checkmark$ & $\checkmark$
& $\mathbf{0.8682}$ & $\mathbf{0.9358}$ \\
\bottomrule
\end{tabular*}
\caption{Component ablation on the private dataset. Stu. = image-to-text prior student and Anc. = modality anchor branch. Prior abbreviations: Rand. = random text prior, Shuf. = shuffled report prior, Fix. = fixed report prior, and Rep. = normal report semantics.}
\label{tab:component_ablation}
\end{table}

Random text and shuffled report controls remain below fixed and student priors, showing that the improvement depends on meaningful normal-report semantics rather than arbitrary or mismatched textual conditions. Student prior further improves over fixed report prior by predicting pseudo-text representations from the current image features, allowing the semantic condition to adapt to each sample under normal-report supervision.

Full ReMoE combines the image-conditioned student prior with the modality anchor branch and achieves the best performance. This result highlights their complementary roles: the student prior provides sample-adaptive normal semantics, while the anchors specify modality and slab directions for RMM, jointly producing more informative feature modulation before encoder-decoder discrepancy computation.

\textbf{Routing strategy.}
In Table~\ref{tab:routing_ablation}, we compare different strategies for combining the routes produced under different modality and slab conditions. Confidence routing assigns larger contributions to routes with stronger predicted responses, while learned routing uses a trainable fusion weight to produce a shared route before expert aggregation.

Response fusion independently computes a modality and slab conditioned expert mixture and combines the resulting responses through concatenation and projection. Response mean directly averages these expert mixtures without introducing an additional fusion module. Its superior performance indicates that direct response aggregation more effectively integrates the contributions of different modality and slab conditions.

\begin{table}[t]
\centering
\small
\setlength{\tabcolsep}{3.5pt}
\begin{tabular*}{\columnwidth}{@{\extracolsep{\fill}}lcc@{}}
\toprule
Strategy & I-AUROC & I-AP \\
\midrule
Confidence routing
& $0.8457\pm0.0031$
& $0.9236\pm0.0024$ \\
Learned routing
& $0.8531\pm0.0024$
& $0.9277\pm0.0019$ \\
Response fusion
& $0.8602\pm0.0022$
& $0.9316\pm0.0017$ \\
Response mean
& $\mathbf{0.8682\pm0.0033}$
& $\mathbf{0.9358\pm0.0020}$ \\
\bottomrule
\end{tabular*}
\caption{Ablation of RMM routing strategies on the private OCT/OCTA dataset. Results are reported as mean $\pm$ standard deviation over five random seeds.}
\label{tab:routing_ablation}
\end{table}

\textbf{Modality combination.}
Table~\ref{tab:modality_combination} analyzes OCT/OCTA input combinations by summarizing single-input results, two-input OCT/OCTA pairs, average three-input combinations, and the full four-input setting, with complete combination results provided in the Supplementary Material. Among the single-input variants, the deep-slab inputs perform better in I-AUROC, with deep OCTA ranking first and deep OCT second, whereas I-AP favors deep OCT and superficial OCTA. This metric-dependent pattern highlights the complementary normality cues captured by different modalities and slabs.

As more inputs are included, performance improves from paired two-input settings to three-input combinations and reaches the best result with all four inputs. This shows that ReMoE benefits from matching report semantics with complete multimodal visual context: report-guided priors provide normal semantic references, while the four OCT/OCTA inputs provide corresponding evidence for feature modulation.

\begin{table}[h]
\centering
\small
\setlength{\tabcolsep}{4pt}
\begin{tabular}{lccc}
\toprule
Modalities & \#Inputs & I-AUROC & I-AP \\
\midrule
OCT-S & 1 & $0.7431$ & $0.8725$ \\
OCT-D & 1 & $0.7822$ & $0.9006$ \\
OCTA-S & 1 & $0.7709$ & $0.8971$ \\
OCTA-D & 1 & $0.7872$ & $0.8761$ \\
Avg. OCT/OCTA pairs & 2 & $0.7994$ & $0.8908$ \\
Avg. 3 inputs & 3 & $0.8195$ & $0.9062$ \\
All inputs & 4 & $\mathbf{0.8682}$ & $\mathbf{0.9358}$ \\
\bottomrule
\end{tabular}
\caption{Modality combination analysis on the private OCT/OCTA dataset. S = superficial slab, D = deep slab.}
\label{tab:modality_combination}
\end{table}

\begin{figure}[t]
    \centering
    \includegraphics[width=0.8\columnwidth]{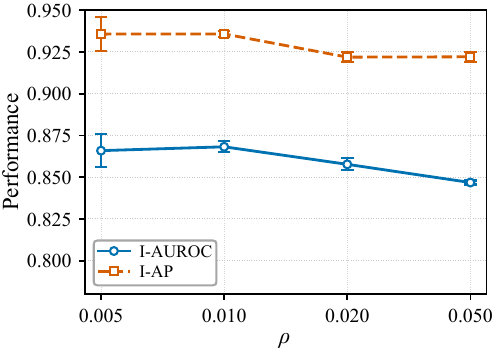}
    \caption{Sensitivity of the top-pixel ratio $\rho$ used for image-level scoring on the private OCT/OCTA dataset.}
    \label{fig:rho}
\end{figure}

\textbf{Hyperparameter sensitivity.}
In Figure~\ref{fig:rho}, we analyze the effect of the top-pixel ratio $\rho$ used for image-level scoring. Performance remains stable at small values and reaches the best result at $\rho=0.01$. Increasing $\rho$ to $0.02$ and $0.05$ leads to an overall performance decline, indicating that averaging a broader set of lower-response pixels may dilute localized anomaly evidence. The similar results at $\rho=0.005$ and $0.01$ further demonstrate stable performance around the selected setting.

In Table~\ref{tab:scale_init_ablation}, we analyze the initialization of the learnable scale parameters $\alpha^q$ in Eq.~\eqref{eq:prior_fusion}. Initializing $\alpha^q$ to 0 achieves the best mean performance, as it suppresses the image-conditioned semantic component only at the beginning of training while allowing its contribution to be progressively adjusted through joint optimization.

\begin{table}[h]
\centering
\small
\setlength{\tabcolsep}{6pt}
\begin{tabular}{lcc}
\toprule
Scale init & I-AUROC & I-AP \\
\midrule
$0$ & $\mathbf{0.8682\pm0.0033}$ & $\mathbf{0.9358\pm0.0020}$ \\
$0.1$ & $0.8526\pm0.0088$ & $0.9271\pm0.0050$ \\
$0.2$ & $0.8573\pm0.0071$ & $0.9298\pm0.0043$ \\
$0.5$ & $0.8624\pm0.0038$ & $0.9325\pm0.0026$ \\
$1.0$ & $0.8580\pm0.0038$ & $0.9296\pm0.0029$ \\
\bottomrule
\end{tabular}
\caption{Ablation of image-conditioned semantic scale initialization on the private OCT/OCTA dataset.}
\label{tab:scale_init_ablation}
\end{table}

\section{Conclusion}

ReMoE provides a Report-Guided Mixture-of-Experts framework for multimodal anomaly detection. It distills normal report semantics into image-conditioned pseudo-text representations and uses modality-aware priors to modulate encoder-decoder feature discrepancies. Experiments on private OCT/OCTA and public OCTA500-3MM settings show consistent improvements over existing visual and multimodal anomaly detection baselines. The current evaluation focuses on retinal OCT/OCTA, and future work will extend ReMoE to broader multimodal medical settings with diverse modality combinations and report styles.

\section*{Ethical Statement}

This study was approved by an institutional ethics committee.
The name of the approving institution and the approval number
are withheld for anonymous review. All procedures were conducted
in accordance with the Declaration of Helsinki, and written
informed consent was obtained from all participants.

\bibliography{aaai2027}


\end{document}